# AnomalyCD: A benchmark for Earth anomaly change detection with high-resolution and time-series observations

Jingtao Li[a], Qian Zhu[b], Xinyu Wang[b*], Hengwei Zhao[a], Yanfei Zhong[a]

[a] State Key Laboratory of Information Engineering in Surveying, Mapping and Remote Sensing, Wuhan University, P. R. China.
[b] School of Remote Sensing and Information Engineering, Wuhan University, P. R. China.
*Corresponding author. Tel.: +86-27-87869685. E-mail address: wangxinyu@whu.edu.cn.

**Abstract**: Various Earth anomalies have destroyed the stable, balanced state, resulting in fatalities and serious destruction of property. With the advantages of large-scale and precise observation, high-resolution remote sensing images have been widely used for anomaly monitoring and localization. Powered by the deep representation, the existing methods have achieved remarkable advances, primarily in classification and change detection techniques. However, labeled samples are difficult to acquire due to the low probability of anomaly occurrence, and the trained models are limited to fixed anomaly categories, which hinders the application for anomalies with few samples or unknown anomalies. In this paper, to tackle this problem, we propose the anomaly change detection (AnomalyCD) technique, which accepts time-series observations and learns to identify anomalous changes by learning from the historical normal change pattern. Compared to the existing techniques, AnomalyCD processes an unfixed number of time steps and can localize the various anomalies in a unified manner, without human supervision. To benchmark AnomalyCD, we constructed a high-resolution dataset with time-series images dedicated to various Earth anomalies (the AnomalyCDD dataset). AnomalyCDD contains high-resolution (0.15–2.39 m/pixel), time-series (3–7 time steps), and large-scale images (1927.93 km$^2$ in total) collected globally. Furthermore, we developed a zero-shot baseline model (AnomalyCDM), which implements the AnomalyCD technique by extracting a general representation from the segment anything model (SAM) and conducting temporal comparison to distinguish the anomalous changes from normal changes. AnomalyCDM is designed as a two-stage workflow to enhance the efficiency, and has the ability to process the unseen images directly, without retraining for each scene. Compared to the traditional methods, AnomalyCDM can suppress the normal changes and increase the recall rate by 10 points. The large-scale AnomalyCDD dataset fully proves the superiority of AnomalyCDM in efficiency and generalization.

1. **Introduction**

Monitoring Earth surface anomalies has long been a critical priority for nations worldwide. Deviating from the historical law and stable state (Wei et al., 2023), anomaly events can develop into natural or man-made disasters and result in fatalities and serious destruction of property (Merz et al., 2021; Zheng et al., 2021). Statistical analysis has shown that anomaly events caused approximately $202.66 billion in economic losses and affected about 0.11 billion people worldwide in 2023 (Coly et al., 2022). To mitigate the harm, the United Nations has set up multi-hazard early warning systems (MHEWS) to protect and save vulnerable communities from destructive anomaly events.

Remote sensing is a common operational tool for anomaly monitoring due to the periodic and large-scale observation ability, which is crucial for the subsequent response (He et al., 2024a). Nations worldwide have established specialized programs for anomaly monitoring and emergency response with Earth observation programs such as the Earth Science Disasters Program in the U.S. supported by the National Aeronautics and Space Administration (NASA) and the Charter program supported by the European Space Agency (ESA). With the development of sensor technology, high spatial resolution remote sensing images are becoming increasingly available, providing more detailed monitoring information. For example, Zheng et al. (2021) processed images from the WorldView-2 and WorldView-3 platforms for precise building damage assessment with six disaster categories, which may not be possible in medium- or low-resolution images.

Modern anomaly detection methods have shown great promise, especially when powered by the deep learning technique, including both classification models and change detection models. However, these models are mostly reliant on human annotation, which makes the trained model limited to certain anomaly categories. Classification models are always trained using single-temporal labeled anomaly samples (He et al., 2024a; Yar et al., 2023), and the change detection models are trained with labeled anomaly samples between the pre-event and post-event images (Zhang et al., 2023). Given a different anomaly category, separate models need to be retrained with category-specific labeled samples, which is not cost-effective (Capliez et al., 2023), since there are various anomaly categories deviating from the normal pattern, as shown in Fig. 1. Furthermore, the existing models struggle to process rare or unseen anomalies with few samples (Tupper and Fearnley, 2023).

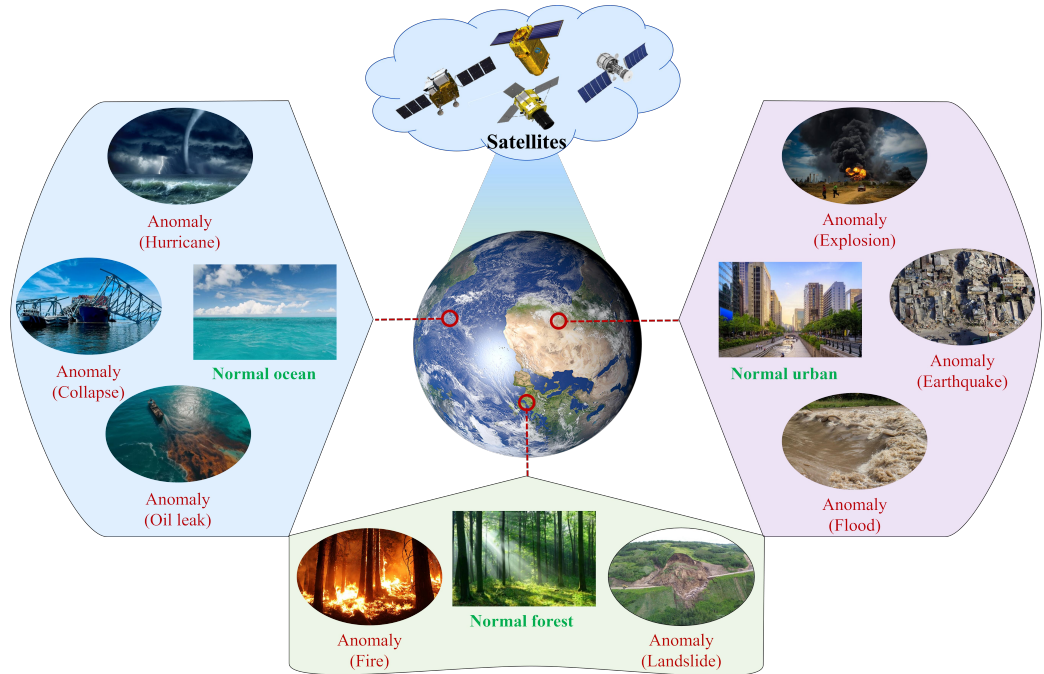

**Fig. 1.** Various Earth surface anomalies deviate from the historical normal pattern in ocean, urban, and forest scenes. Designing a unified anomaly detection model is necessary, considering the cost efficiency and generalizability for unseen or rare anomalies.

To overcome this barrier, we propose the anomaly change detection (AnomalyCD) technique, which can accept an unfixed number of time steps and localize anomalies from the perspective of the dynamic change pattern. Our core observation is that normal changes are constantly occurring, while an anomaly corresponds to a change that has not occurred before. Assuming that the changes in historical images are normal due to the low probability of occurrence (Khandelwal et al., 2017), continuous observations can help us to distinguish between anomalous changes and normal changes, where the normal changes are commonly caused by periodic activities and varying imaging conditions. Compared to the existing techniques shown in Fig. 2, AnomalyCD relies on dynamic observations to filter out the anomalous changes without human supervision, where the trained model has the potential to detect various anomalies in a unified manner.

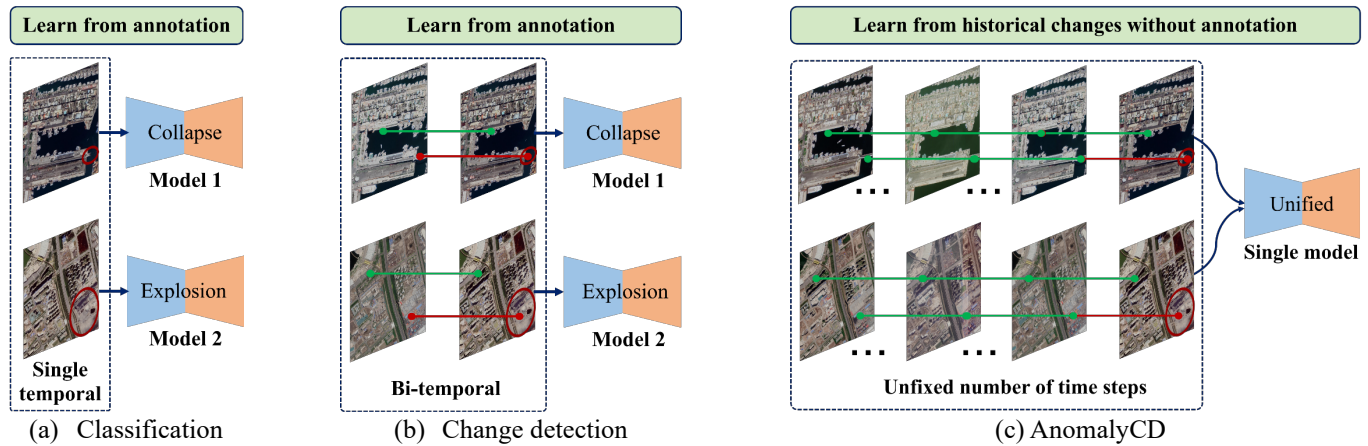

**Fig. 2.** Comparison between the proposed AnomalyCD technique and the existing classification and change detection techniques. (i) AnomalyCD accepts an unfixed number of time steps and learns from the historical images without human supervision, where the historical changes are assumed to be normal (green line) and any unseen change (red line) is treated as an anomaly. In contrast, classification and change detection techniques deal with fixed single- or bi-temporal images and rely on human annotation to complete the training process. (ii) Without the constraint from labeled samples, AnomalyCD can detect various anomaly categories in a unified manner, while the classification and change detection models need to be retrained given a different anomaly category.

To benchmark the AnomalyCD technique, we first built a high-resolution dataset with time-series images for the Earth anomaly detection task. The anomaly change detection dataset (AnomalyCDD) is characterized by three aspects, i.e., the targeted Earth anomaly detection task, time-series images, and large-scale observations. Differing from the traditional datasets dedicated to general changes or building changes, AnomalyCDD focuses on six categories of Earth anomalies that have caused serious damage, which were all collected from news events. For the pre-event images, all the available images over the nearest 3 years were collected to form the historical time-series observations. Most events have 4–5 time steps for extracting the normal changes from the time-series images, and some events have 6–7 steps. Precise pixel-level annotation is provided for each event, including both the anomaly location and damage category. With the high spatial resolution (0.15–2.39 m/pixel), most of the images are over 10000 × 10000 pixels in size, covering a total of 1927.93 km$^2$. The collected anomaly events are dispersedly distributed at a global scale.

Based on the AnomalyCDD dataset, a zero-shot anomaly change detection model (AnomalyCDM) was further developed via the temporal visual representation comparison extracted from the segment anything model (SAM) (Kirillov et al., 2023). Following the ethos of the AnomalyCD technique in Fig. 2, AnomalyCDM is designed as a two-stage workflow, where the first stage detects all the change locations as anomaly candidates, using only the latest bi-temporal images, and the second stage further filters out the normal changes. The change score is computed by comparing the corresponding SAM embeddings from the different time steps, and the changes that happened in the historical images are all considered as normal in the second stage. The SAM was chosen considering the proven property that extracted embeddings can be semantically grouped for unseen images, which enhances the zero-shot detection ability. Fig. 3 shows the zero-shot performance, referring to the anomaly categories of collapse, obstruction, and explosion, where AnomalyCDM is able to process the images directly, without retraining. Given the change map of the first stage, AnomalyCDM examines all the historical images (first column) and realizes that the changing ships (Fig. 3a) and fluvial movement (Fig. 3b) belong to the normal changes caused by periodic activities. A similar situation can be observed in Fig. 3c for the normal changes caused by the imaging differences of the same building. AnomalyCDM achieves an average anomaly event detection F1-score of 55.62, surpassing the traditional change detection models by a large margin of around 13 points.

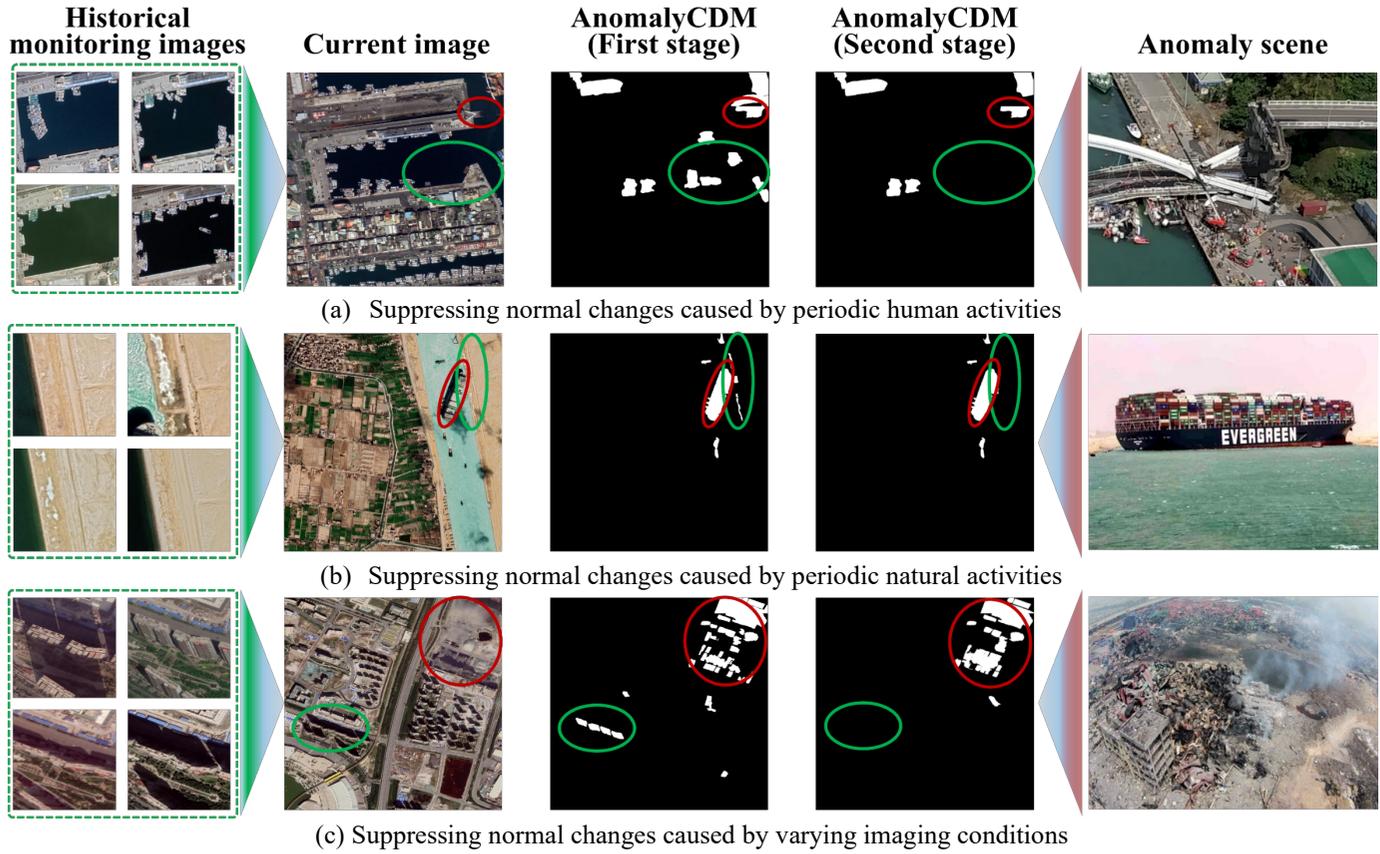

**Fig. 3.** Exemplified normal changes caused by periodic activities and imaging conditions, where the red circles represent real anomaly regions and the green circles represent the normal change regions. From top to bottom, the anomaly events are (a) the collapse of the Nanfang'ao Bridge in Taiwan, China (2019), (b) the obstruction of the Suez Canal (2021), and (c) the warehouse explosion in the Port of Tianjin in China (2015). Periodic activities are caused by human activities (e.g., the harbor ships in (a)) or natural activities (e.g., caused by fluvial movement in (b)), where the first stage in AnomalyCDM detects all the changes, and the second stage suppresses the periodic activities by analyzing all the historical images. A similar performance can be observed for suppressing the normal changes caused by varying imaging conditions (e.g., satellite viewing angles in (c)).

The main contributions of this paper are summarized below:

1. We propose the AnomalyCD technique for Earth surface anomaly detection. Differing from the existing techniques, AnomalyCD learns to distinguish the normal and anomalous changes from an unfixed number of historical time steps. Without the limitation of human supervision, the trained model of AnomalyCD can process varied anomalies in a unified manner.

2. An anomaly change detection dataset—AnomalyCDD—was constructed, dedicated to the Earth anomaly detection task. AnomalyCDD contains high-resolution (0.15–2.39 m/pixel), time-series (3–7 time steps), and large-scale images (1927.93 km$^2$ in total) collected from 80 anomaly events globally.

3. An anomaly change detection baseline model—AnomalyCDM—was designed, which can detect the unseen images in a zero-shot manner with the temporal visual representation from the SAM latent space. The two-stage workflow helps AnomalyCDM identify the real anomalous changes effectively.

The rest of this paper is organized as follows. Section 2 introduces the related work about Earth anomaly detection. Section 3 presents the constructed AnomalyCDD dataset and the detailed statistical information. The proposed AnomalyCDM baseline is presented in Section 4. Section 5 validates the zero-shot detection performance of AnomalyCDM on the constructed dataset. Finally, the paper is concluded in Section 6.

## 2. Related work

### 2.1 Earth surface anomaly detection in the deep learning era

Over the past decade, the performance of Earth surface anomaly detection has made significant advances across multiple anomaly categories, propelled by deep learning (Wei et al., 2023). The main anomaly detection methods include classification models (He et al., 2024a; Yar et al., 2023) and change detection models (Zhang et al., 2023; Zheng et al., 2021). Classification models rely on single-temporal labeled samples and are always trained for specific anomaly events (including both image level and pixel level). For example, Yar et al. (2023) trained a fire detection model with 3000 labeled normal and fire images from drone and satellite platforms, where diverse scenes, including night-time and cloudy weather conditions, were collected to enhance the model robustness. He et al. (2024b) focused on the anomaly type of flood, and trained a segmentation model with 2862 fully labeled samples from the high-resolution Calgary-Flood dataset. However, under supervised learning, the scale and quality of the labeled anomaly samples has a great influence on the model performance (Pang et al., 2021).

Differing from classification models, change detection models utilize bi-temporal images for anomaly localization, including pre-event and post-event images. To reduce the label cost, change detection models for anomaly detection mostly belong to the branch of binary change detection, which includes only two annotated categories of anomalous changes and no changes. For instance, Zhang et al. (2023) used 630 image pairs with binary labels to monitor the large-scale anomaly of a landslide in Japan, using source high-resolution images from the SPOT 6 satellite. In addition to localizing anomalies, some studies have further categorized fine-grained levels. For example, Zheng et al. (2021) used the large-scale xBD dataset to not only localize damaged buildings but also recognize four damage levels, such as minor damage or destroyed. Benefiting from the reference of pre-event images, change detection models always have fewer false alarms and dominate the Earth anomaly detection task.

The success of change detection models is evident across various anomaly types. Despite this, the models are mostly supervised and trained for fixed anomaly categories, which hinders their application for anomalies with few samples or unknown anomalies. Although some unsupervised (Wu et al., 2023) or zero-shot change detection models (Zheng et al., 2024) have less need for sample annotation, they detect all kinds of changes, without the supervision constraints, including the normal changes, as shown in Fig. 3, and bring more false alarms.

*2.2 Change detection with time-series observations (more than two)*

The proposed AnomalyCD method uses an unfixed number of time steps to detect anomalous changes, so it is necessary to review the related techniques for time-series observations, which include continuous change detection (Zhu and Woodcock, 2014) and abrupt change detection (Brakhasi et al., 2021). Continuous change detection was first proposed by Zhu and Woodcock in 2014, where it was used for the dynamic land-cover mapping task with all the available Landsat observations (Zhu and Woodcock, 2014). The continuous change detection and classification (CCDC) algorithm was adopted by the United States Geological Survey (USGS) as the official algorithm for monitoring land-cover changes (Xian et al., 2022). The CCDC algorithm detects a real change first and then updates the corresponding land-cover types, where a real change means that the corresponding pixel is observed to change in multiple consecutive images. Differing from the CCDC algorithm, abrupt change detection uses historical observation data to fit a predictor, where a large prediction error represents abrupt change. The abrupt change detection technique is mostly used for point data with statistical trends. For example, Li et al. (2022) used 20-year MODIS long-term land surface temperature (LST) data and the abrupt change detection technique to detect abrupt changes in an LST time series. Abera et al. (2022) used multi-source data from satellite observations (Landsat and MODIS) and airborne laser scanning data to monitor the net fractional woody cover abrupt change during 2001–2019 over Ethiopia and Kenya

Compared to continuous change detection and abrupt change detection, the proposed AnomalyCD technique also uses images from multiple time steps, but differs in the detection principle. Serving dynamic mapping tasks, continuous change detection uses multiple post-event time steps to ensure that a land-cover change happened and needs to be updated. In contrast, AnomalyCD uses multiple historical images (i.e., pre-event) to discriminate the anomalous changes, which can achieve a faster anomaly response than continuous change detection. Abrupt change detection focuses on point observations and models the statistical trends, where a large prediction error represents an abrupt change. Focusing on image data, AnomalyCD defines an anomalous change as a change that has never occurred in historical data, where the trained model can also be aware of the normal changes that do not follow a statistical trend.

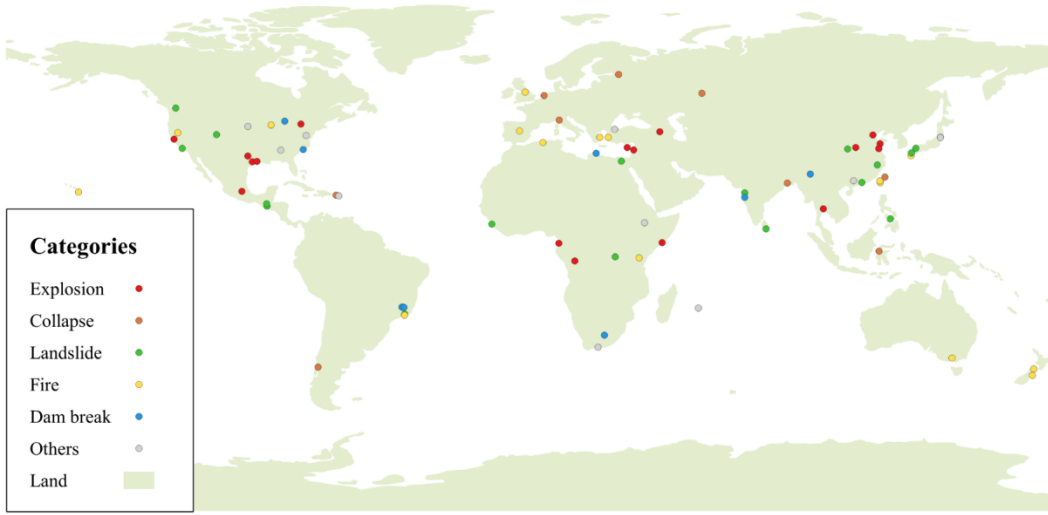

**Fig. 4.** The global spatial distribution of the collected anomaly events, where the collected events cover different states and countries.

3. **AnomalyCDD: A dataset for anomaly change detection**

To benchmark the AnomalyCD technique, we built a new dataset called AnomalyCDD with high-resolution time-series images covering various Earth anomaly categories. In this section, we first give an overview of the constructed dataset, followed by some necessary statistics. Based on this, the unique characteristics of the AnomalyCDD dataset are highlighted by comparing it with the existing related datasets.

*3.1  Dataset overview*

The AnomalyCDD dataset has six categories, covering 80 large-scale anomaly events worldwide with time-series observations. Table 1 and Fig. 4 give an overview description of the dataset. The anomaly categories in Table 1 refer to both natural (e.g., landslide) and man-made anomaly events (e.g., large-scale explosion). Some anomaly categories have only a few events, such as plane crashes and tornados, which are categorized as "others". Fig. 4 shows the spatial distribution of the collected anomaly events, where the events of each category are distributed in different states and nations. Each anomaly event is related to large-scale time-series images, where the total covered area reaches 1927.93 km$^2$. The various anomaly categories, the globally dispersed distribution, and the large-scale observations make the dataset very challenging, especially without any human supervision.

We collected the anomaly events from news reports and then downloaded the corresponding time-series images from the Google Earth platform. For the pre-event images, all the available images in the nearest 3 years were collected to form the time-series observations. Some event locations can have long satellite revisit periods, so we extended the collection range from 3 years to 8 years for these events. Most of the images have a spatial resolution in the range of 0.15–2.39 m/pixel. The anomaly events are labeled at the pixel level, according to the event location (longitude and latitude) and the related news reports.

**Table 1**

Anomaly categories in the AnomalyCDD dataset, which includes six categories and covers a large-scale region.

| Category | Description | Anomaly events | Coverage area (km²) |
|---|---|---|---|
| Explosion | A sudden and violent release of energy, often resulting in the generation of high temperatures, gases, and pressure, which cause damage to structures and injuries to living beings. | 18 | 194.54 |
| Collapse | The abrupt structural collapse of a building, infrastructure, or any constructed entity, caused by the compromise of its structural integrity or poor construction quality over time. | 9 | 49.25 |
| Landslide | A geologic event characterized by the mass movement of soil, rocks, and debris down a slope, typically triggered by heavy rainfall, earthquakes, or human activities. | 17 | 484.30 |
| Fire | An unexpected and uncontrolled fire that spreads rapidly, including building fires, forest fires, and volcanic eruptions. It often results in building collapse or widespread damage. | 16 | 255.18 |
| Dam break | Dam breaks can result from factors such as overtopping, foundation instability, or design flaws, leading to the unrestricted release of stored water or other materials. | 9 | 652.59 |
| Others | This category encompasses events that cannot be specifically classified into the categories described above, such as plane crashes, oil spills, terrorist attacks, tornadoes, earthquakes, and so on. | 11 | 322.06 |

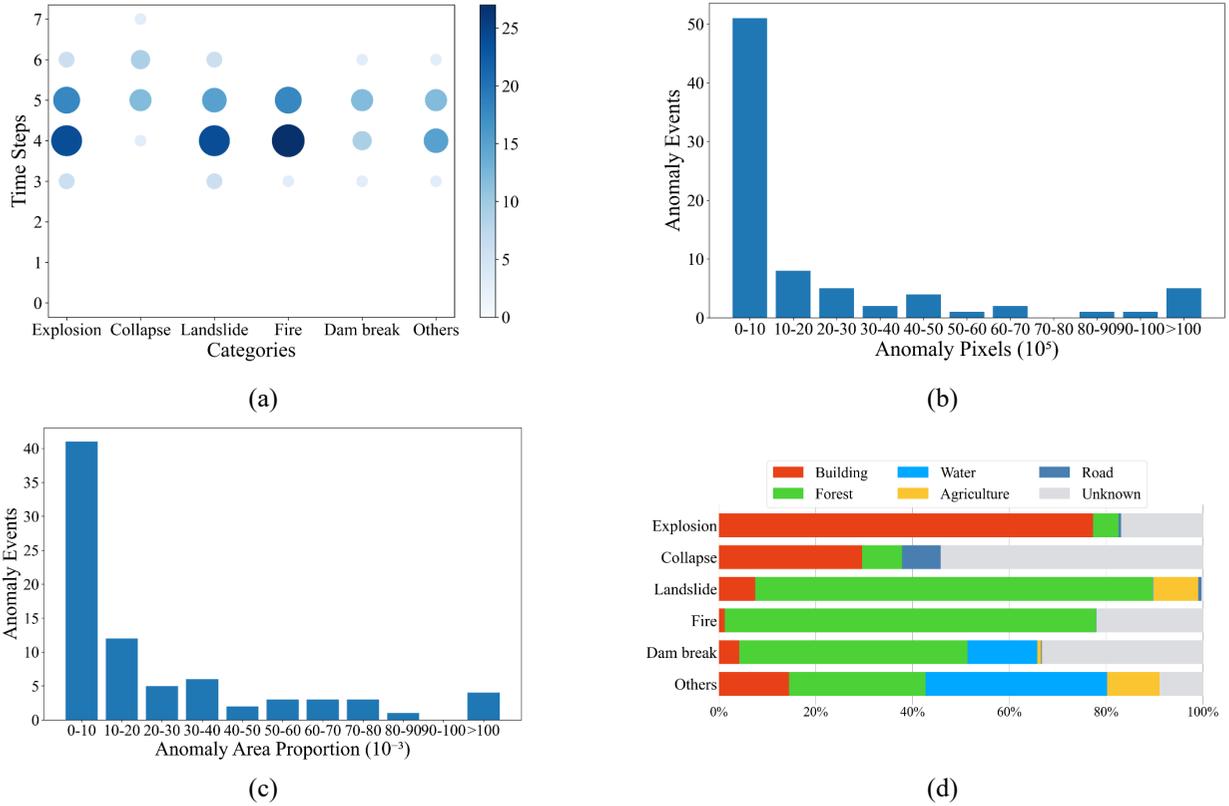

**Fig. 5.** Statistical information for the AnomalyCDD dataset, including (a) the time step number, (b) the anomaly pixel number, (c) the anomaly area proportion, and (d) the land-cover categories of the anomaly regions.

## 3.2 Statistics for AnomalyCDD

To give a comprehensive description, statistics are provided in Fig. 5 about the time steps (Fig. 5a), anomaly pixel number (Fig. 5b), anomaly area proportion (Fig. 5c), and the land-cover categories of the anomaly regions (Fig. 5d). In Fig. 5a, most events have 4–5 time steps, but some have 6–7 time steps. The anomaly category of collapse has the largest number of time steps (mostly 5–7 time steps), which may be because most such events occur in urban areas with shorter revisit periods. Fig. 5b and Fig. 5c show the anomaly pixel number and area proportion of each event. Most events have anomaly regions with several hundred meters width and height, occupying a proportion of around 1–3% in the collected large-scale images. In addition to the anomaly location, we have also labeled the affected land-cover categories for each anomaly event, as shown in Fig. 5d, to support the damage assessment requirement. Six classical land-cover categories are considered, where the "others" category refers to the pixels that do not belong to the previous five categories.

## 3.3 Characteristics of AnomalyCDD

Compared to the existing high-resolution change detection datasets (Table 2), AnomalyCDD has three unique characteristics: the various Earth anomaly categories, time-series images, and large-scale observations.

(1) **Various Earth anomaly categories**. Most datasets are built for general change monitoring (e.g., from grass to building) (Lebedev et al., 2018) or only for building changes (e.g., the LEVIR-CD (Chen and Shi, 2020) and WHU Building (Ji et al., 2018) datasets). In contrast, the AnomalyCDD dataset is expected to facilitate the detection of Earth anomalies, where the collected events refer to six anomaly categories, as listed in Table 1.

(2) **Time-series images**. The current high-resolution change detection studies mostly concentrate on bi-temporal images, while the proposed AnomalyCDD dataset has 4–5 time steps for most of the anomaly events. The time-series images were collected to help the AnomalyCD model distinguish the anomalous changes from the normal changes. The DynamicEarthNet dataset is a special case, which has 730 daily time steps from the Planet platform (Toker et al., 2022). However, it focuses on general changes and is not suitable for the Earth anomaly detection task.

(3) **Large-scale observations**. AnomalyCDD encompasses 80 anomaly events covering 1927.93 km$^2$ in

**Table 2**
Statistical comparison between AnomalyCDD and the existing change detection datasets. AnomalyCDD focuses on the changes related to Earth anomalies, and the images are large-scale time-series images with a high spatial resolution.

| Dataset | Resolution (m) | Source | Area (km$^2$) | Time steps | Image size | Image number | Focus |
|---|---|---|---|---|---|---|---|
| LEVIR-CD (Chen and Shi, 2020) | 0.5 | Google Earth | 166.96 | 2 | 1024 | 637 | Changes related to buildings |
| CDD (Lebedev et al., 2018) | 0.03–1.0 | Google Earth | \ | 2 | 256 | 16000 | General changes |
| WHU Building (Ji et al., 2018) | 0.075, 0.3–2.5, 2.7 | QuickBird, Worldview series, IKONOS, and ZY-3 | \ | 2 | 512 | 25679 | Changes related to buildings |
| SECOND (Yang et al., 2021) | 0.5–3.0 | \ | \ | 2 | 512 | 4662 | General changes |
| Hi-UCD (Tian et al., 2020) | 0.1 | Estonian Land Board | 13.56 | 3 | 1024 | 1293 | Urban changes |
| DynamicEarthNet (Toker et al., 2022) | 3.0 | PlanetFusion | 707.78 | 730 | 1024 | 27375 | General changes |
| AnomalyCDD | 0.15–2.39 | Google Earth | 1927.93 | 3–7 | 1872–22463 | 80 | Changes related to Earth anomalies |

total, which greatly surpasses the other datasets listed in Table 2. With the high spatial resolution, most of the images are over 10000 × 10000 pixels in size. Furthermore, the anomaly events are dispersedly distributed at a global scale. The large-scale setting in area and spatial distribution can not only reflect the advantage of the remote sensing technique for the Earth anomaly detection task, but also validates the model generalizability since an overfitting deep model would perform poorly in large-scale scenes.

## 4. AnomalyCDM: Anomaly change detection with time-series images

The AnomalyCD technique represents an unsupervised and unified solution for the Earth anomaly detection task. To instantiate it and set a baseline for the AnomalyCDD benchmark, a zero-shot anomaly detection model named AnomalyCDM was developed, which can process an unfixed number of time steps and localize various anomalies without human supervision. Based on the SAM (Kirillov et al., 2023), AnomalyCDM does not need fine-tuning for a new data distribution and can give a rapid response in practical usage. In the following, we first give an overview of AnomalyCDM and then go into the details of the model.

### 4.1 Model overview

The framework of AnomalyCDM is shown in Fig. 6, which is a two-stage workflow of first detecting all the changes and then distinguishing the anomalous changes. Due to the low probability property of anomaly events, the historical images can be seen as anomaly-free, where the changes belong to the normal state, which provides the basis for distinguishing the anomalous changes in the second stage.

In the first stage, the nearest bi-temporal images are used to detect all the changes. The time-series observations with $n$ historical images are denoted as $\{\mathbf{T}_n,...,\mathbf{T}_2,\mathbf{T}_1,\mathbf{X}\}$, where $\mathbf{X}$ is the current image to be detected and $\mathbf{T}_1 - \mathbf{T}_n$ are the historical images assumed to be normal. $\mathbf{T}_1$ comes from the nearest time step and $\mathbf{T}_n$ from the farthest. Although changes may not be anomalies, anomalies must appear as change locations between $\mathbf{X}$ and $\mathbf{T}_1$ in turn. Thus, the first stage can be considered as selecting the anomaly candidates for the later processing, which can reduce the processing burden of the second stage significantly since most areas are unchanged. We denote the proposed module responsible for the first stage as the SAM bi-temporal change detection (SAM Bi-CD) module.

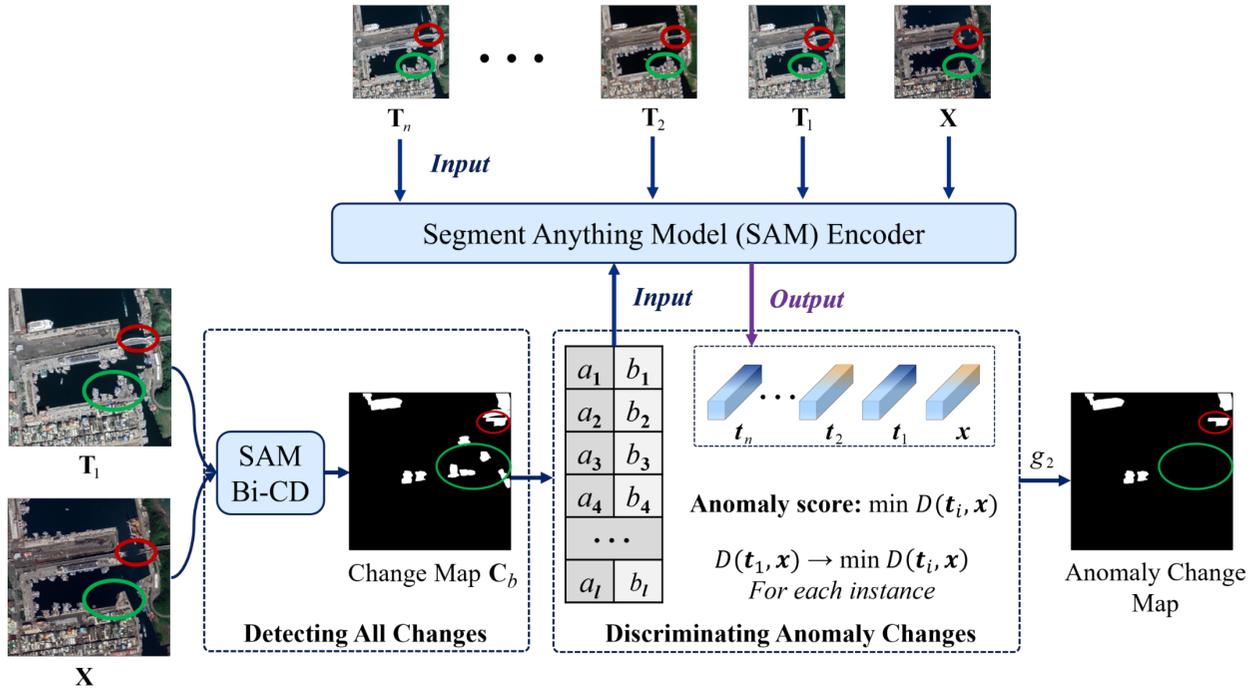

**Fig. 6.** The overall workflow of AnomalyCDM. AnomalyCDM has two stages, where the first stage detects all the changes as anomaly candidates with the designed SAM Bi-CD module, and the second stage discriminates the anomalous changes further by comparing the temporal representations.

At the second stage, the detected changes are processed further to distinguish the real anomalous changes with the time-series images. The process is conducted at the instance level, referring to the SAM segmentation masks. For each instance mask, the corresponding $n+1$ embeddings are cropped from the time-series images, which includes the embedding $x$ for $\mathbf{X}$ and $t_1 - t_n$ for $\mathbf{T}_1 - \mathbf{T}_n$. Assuming that $t_1 - t_n$ represents the normal pattern, once there is a $t_i$ similar to $x$, the mask would obtain a small anomaly score, even if $t_1$ and $x$ have a great difference. Since normal changes are mostly caused by periodic activity or the changing imaging conditions and have occurred before, this operation can suppress the normal changes effectively and retain only the real anomalies.

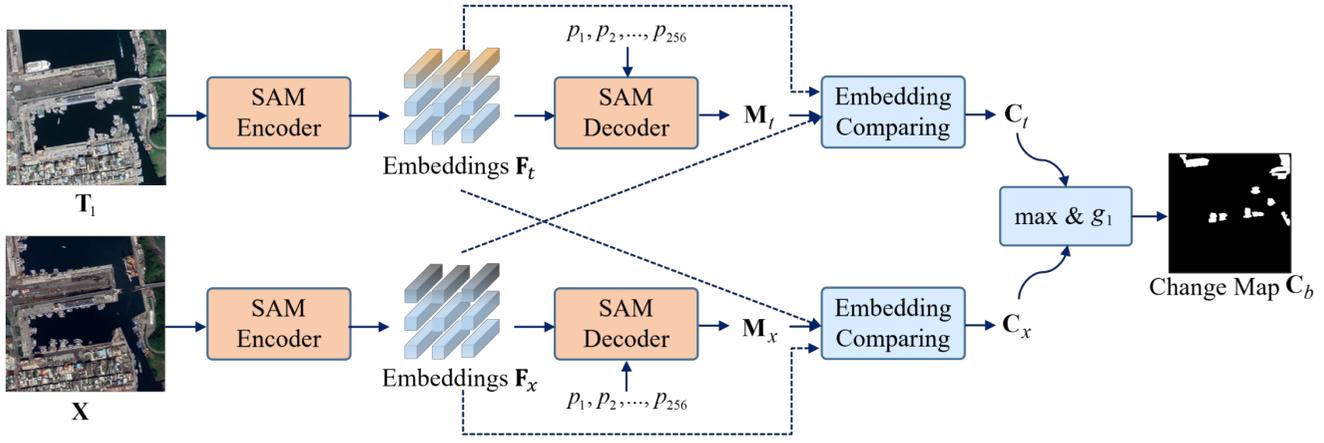

**Fig. 7.** The detailed workflow of the SAM Bi-CD module in the first stage, which conducts bi-temporal change detection to filter out the anomaly candidates. To deal with the two cases of object disappearance and appearance simultaneously, we compare the instance embeddings twice for both $\mathbf{M}_x$ and $\mathbf{M}_t$. The comparison is conducted in the embedding space to reduce the impact of geometric registration error.

*4.2 Stage 1: Detecting all changes*

SAM Bi-CD detects all the changes between $\mathbf{X}$ and $\mathbf{T}_1$ as anomaly candidates at this stage since the anomalies will be part of the change locations. Fig. 7 shows the detailed bi-temporal change detection workflow, where the SAM is used twice for the bi-temporal images. With the prompts of grid points $p_1, p_2, ..., p_{256}$, the embeddings in the feature space (i.e., $\mathbf{F}_t$ and $\mathbf{F}_x$) and the corresponding segmentation masks (i.e., $\mathbf{M}_t$ and $\mathbf{M}_x$) are extracted separately. With $\mathbf{M}_t$ and $\mathbf{M}_x$, prior works have attempted to compare the mask shapes directly and treated a low intersection over union (IoU) value as a high change probability (Chen et al., 2024a). However, we found that a small geometric registration error can bring serious disturbance, so we chose to compare the masks in the embedding space. For a binary instance mask $\mathbf{M}_{t,j} \in \mathbf{M}_t$, the corresponding mask embeddings are cropped from $\mathbf{F}_t$ and $\mathbf{F}_x$ (e.g., $\mathbf{F}_t(\mathbf{M}_{t,j}==1)$, $\mathbf{F}_x(\mathbf{M}_{t,j}==1)$) and then their mean embeddings are compared with the distance metric $D$. The final obtained change score $S(\mathbf{M}_{t,j})$ is given as follows:

$$S(\mathbf{M}_{t,j}) = D(\mathbf{F}_t(\mathbf{M}_{t,j}==1), \mathbf{F}_x(\mathbf{M}_{t,j}==1)) \quad (1)$$

To deal with the two cases of object disappearance and appearance simultaneously, the embedding comparison operation is applied for both segmentation masks $\mathbf{M}_x$ and $\mathbf{M}_t$. The mask is segmented in $\mathbf{M}_x$ for the object appearance case and in $\mathbf{M}_t$ for the object disappearance case. After the comparison in both directions, two change density maps (i.e., $\mathbf{C}_t$ and $\mathbf{C}_x$) can be obtained with the same spatial shape. To ensure that all the changes are detected, the maximum value from $\mathbf{C}_t$ and $\mathbf{C}_x$ for each overlapped mask is taken and a threshold binarization operation (i.e., $g_1$) is applied. The final output binary change map $\mathbf{C}_b$ can be represented as follows:

$$\mathbf{C}_b = g_1(\max(\mathbf{C}_t, \mathbf{C}_x)) \qquad (2)$$

*4.3  Stage 2: Discriminating anomalous changes*

Based on the anomaly candidates in $\mathbf{C}_b$, all the time-series images are introduced in this stage to ensure that the model is aware of the normal and anomalous changes. The core idea is to treat all the historical images $\mathbf{T}_1 - \mathbf{T}_n$ as normal, and the changed instances in $\mathbf{C}_b$ are considered normal if the instance pattern has been observed in any time step of $\mathbf{T}_1 - \mathbf{T}_n$. By considering all the time steps, the normal changes caused by periodic activity and the change of imaging conditions can be largely suppressed.

Similarly, we conducted a comparison in the embedding space of all the time steps to deal with the registration error for each instance, as shown in Fig. 6. Specifically, the images of all the time steps were converted into the embedding space with the SAM encoder. Given an instance mask $\mathbf{M}_k$, the corresponding embedding from all the time steps was cropped according to the coordinates $(a_1, b_1) - (a_l, b_l)$. As in stage 1, the mean embedding vector was taken to represent the mask, and the obtained embeddings were denoted as $x, t_1, t_2, ..., t_n$. Differing from the change score $S$ computed with two time steps, the anomaly change score $S_a$ compares all the time steps together, as follows:

$$S_a(\mathbf{M}_k) = \min(\sum_{i=1}^{n} D(\mathbf{x}, \mathbf{t}_i)) \qquad (3)$$

which assign a continuous anomaly score for each mask. By conducting a threshold binarization operation again (i.e., $g_2$), the binary anomaly change map can be finally obtained.

## 5. Experiments and analysis

*5.1 Experimental settings*

**Dataset**: The large-scale AnomalyCDD dataset built in this study was used to evaluate the detection models. Due to the large-scale property and the diverse anomaly categories, AnomalyCDD can prevent the overfitting problem and provide a more objective result.

**Evaluation metrics**: The conventional change detection metrics are used in this paper, i.e., recall, precision, and $F_1$-score. The recall reflects how many anomaly pixels have been detected, and the precision reflects how many detected anomaly pixels are correct. Since most anomalous change pixels always occupy an extremely low proportion (less than 4%), and it is more important to find the anomalies than suppress the false alarms, the precision was computed by weighting the true anomalies and false anomalies in a ratio of 1:10. The $F_1$-score balances the recall and precision and acts as a comprehensive metric.

**Comparison methods**: Due to the large-scale observations and various anomaly categories, it was infeasible to apply some of the existing trained models or provide comprehensive training labels for supervised training. Label-free change detection methods were therefore chosen for the comparison analysis, including: (i) unsupervised change detection models (image differencing (ID) (Lipton et al., 1998), change vector analysis (CVA) (Bovolo and Bruzzone, 2006), and fully convolutional change detection with generative adversarial network (FCD-GAN) (Wu et al., 2023); (ii) zero-shot change detection models (SAM-multimodal change detection (SAM MCD) (Chen et al., 2024b) and the SAM Bi-CD module in the proposed AnomalyCDM; and (iii) AnomalyCDM and its variant SAM-change vector analysis (SAM-CVA) model, where we extend the time-series comparison approach into the CVA method with the same SAM latent feature space.

**Implementation details**. Except for FCD-GAN, the remaining models inferred the test images directly,

without training. FCD-GAN was trained with a learning rate of 0.0002, batch size of 8, and 20 training epochs. The remaining settings were consistent with the original papers. Since the label-free methods can only output a continuous change density map, we chose the 0.94 quantile to obtain the binary map, considering the low ratio prior of anomalies (see the related analysis in Section 5.2.4). The SAM-based models (e.g., SAM Bi-CD and AnomalyCDM) processed the large-scale images with a non-overlapped patch size of 2048 pixels (see the related analysis in Section 5.2.5). As in Fig. 6, SAM Bi-CD acted as a module in AnomalyCDM and provided the coarse anomaly candidates, where only the top 30% change instances were considered further with the time-series observations to judge the anomaly degree for efficient processing. We used the SAM-based version to segment the masks, where $16 \times 16$ point prompts were used with an IoU prediction threshold of 0.7 and a stability score threshold of 0.4.

*5.2   Experimental results and analysis*

*5.2.1   Quantitative comparison results*

We report the quantitative results in Table 3, including the results for each anomaly category and the average results. Although ID and CVA can process the images efficiently with simple mathematical calculations, they detect in the original image space and are sensitive to the normal changes caused by the imaging conditions, leading to a poor performance, especially for large-scale scenes. The obtained $F_1$-scores are mostly lower than 0.20. As an unsupervised method, FCD-GAN introduces the generative adversarial network (GAN) to map the relationship of the two time step images, where the deep representation and regression prior enhance the $F_1$-score performance to 0.30. Differing from FCD-GAN, which needs to be retrained in the target domain, SAM MCD and SAM Bi-CD are powered by the remarkable SAM model, where the generalizable segmentation performance supports the zero-shot detection. Both models achieve an average $F_1$-score performance of over 0.40. The proposed SAM Bi-CD module shows a superior performance since we do not compare the segmented masks directly, but instead the latent features, which can suppress the influence of registration error and projection angle difference, especially for buildings.

With the generalizable feature space of the SAM, the proposed AnomalyCDM introduces time-series observations and increases the $F_1$-score performance further by 7 points. Since the comparison models used the same 0.94 quantile, the superior performance of AnomalyCDM proves the effectiveness of time-series information for suppressing the normal changes and increasing the anomaly recall. A similar improvement can be observed in the SAM-CVA model, which shows the robustness of this key insight for different implementations.

*5.2.2 Qualitative comparison results*

Partial and whole detection maps are visualized in Figs. 8 and 9. The partial results are listed for each anomaly category, and the bi-temporal results of FCD-GAN and SAM Bi-CD are also shown. Since FCD-GAN is conducted at the pixel level and SAM Bi-CD at the instance level, the maps of SAM Bi-CD have fewer noisy pixels and are more complete. SAM Bi-CD detects all the changes in the first stage of AnomalyCDM, and the detected results are processed further in the second stage for distinguishing the normal/anomalous changes. The last two columns in Fig. 8 demonstrate the effect of the time-series

**Table 3**

Quantitative comparison results on the constructed AnomalyCDD dataset. The models did not need labeled samples and could process the test images directly.

| | Explosion | | | Collapse | | | Landslide | | | Fire | | | Dam break | | | Others | | | Average | | |
|---|---|---|---|---|---|---|---|---|---|---|---|---|---|---|---|---|---|---|---|---|---|
| | R | P | F1 | R | P | F1 | R | P | F1 | R | P | F1 | R | P | F1 | R | P | F1 | R | P | F1 |
| ID | 3.72 | 2.63 | 3.07 | 9.81 | 4.82 | 6.46 | 4.76 | 1.40 | 2.16 | 5.87 | 5.71 | 5.79 | 0.00 | 0.00 | 0.00 | 15.76 | 10.34 | 12.49 | 6.65 | 4.15 | 5.00 |
| CVA | 11.72 | 6.39 | 8.27 | 33.84 | 22.29 | 26.88 | 22.47 | 18.45 | 20.26 | 18.85 | 18.57 | 18.71 | 10.55 | 7.90 | 9.04 | 22.22 | 11.85 | 15.46 | 19.94 | 14.24 | 16.44 |
| FCD-GAN | 26.59 | 24.57 | 25.54 | 39.85 | 25.80 | 31.32 | 36.81 | 39.27 | 38.00 | 23.28 | 29.43 | 26.00 | 37.40 | 72.35 | 49.31 | 25.82 | 29.29 | 27.45 | 31.63 | 36.79 | 32.94 |
| SAM MCD | 28.61 | 33.26 | 30.76 | 62.34 | **43.98** | 51.58 | 50.72 | 57.28 | 53.81 | 32.96 | 51.32 | 40.14 | 31.28 | 82.14 | 45.30 | 27.10 | **46.86** | 34.34 | 38.84 | 52.47 | 42.66 |
| SAM Bi-CD | 39.38 | 28.03 | 32.75 | 70.68 | 33.09 | 45.07 | 61.87 | 48.10 | 54.12 | 56.15 | 50.47 | 53.16 | 52.67 | 79.73 | 63.43 | 41.31 | 43.52 | 42.39 | 53.68 | 47.16 | 48.49 |
| SAM-CVA | 46.47 | **37.54** | 41.53 | 70.33 | 38.61 | **49.85** | 67.20 | **55.30** | **60.67** | 60.49 | **56.44** | 58.39 | 63.76 | **88.97** | 74.28 | 36.57 | 45.03 | 40.36 | 57.47 | **53.65** | 54.18 |
| AnomalyCDM | **55.68** | 36.30 | **43.95** | **76.03** | 36.20 | 49.04 | **72.51** | 52.07 | 60.61 | **64.96** | 54.26 | **59.13** | **69.77** | 87.68 | **77.71** | **42.43** | 44.14 | **43.27** | **63.56** | 51.78 | **55.62** |

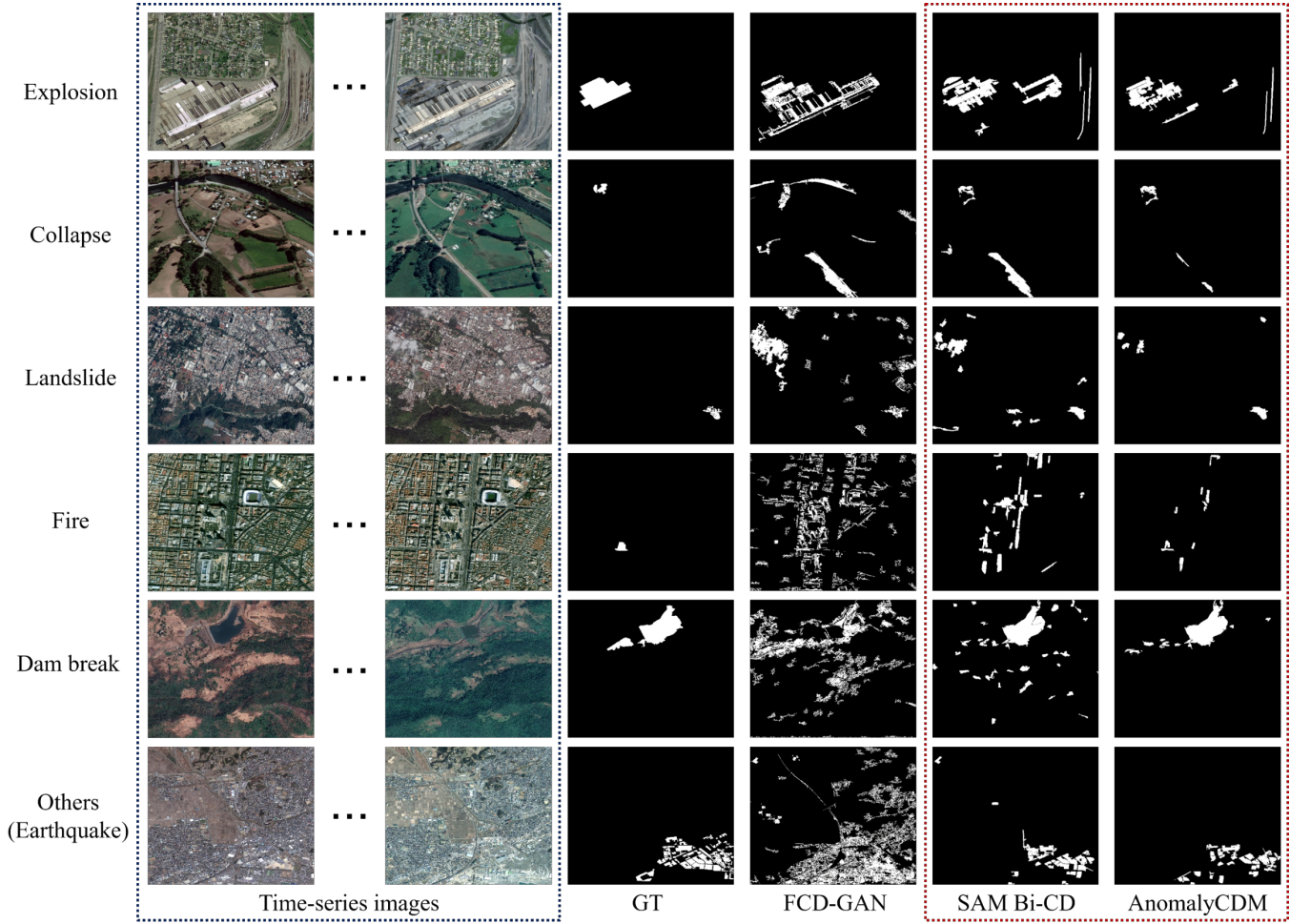

**Fig. 8.** Qualitative comparison results for the six anomaly categories. Compared to the traditional change detection methods, the proposed AnomalyCDM can make the model aware of anomalous changes and reduce the false alarms.

observations for suppressing most normal changes.

The constructed AnomalyCDD dataset covers nearly 2000 km$^2$ in total, and most of the anomaly events are of a large scale. To show the large-scale detection performance, two exemplified anomaly events are shown completely in Fig. 9. The first event corresponds to the landslide in Badul, Sri Lanka (2014), with an image size of $6160 \times 6111$. The second event corresponds to a fire that broke out in West Yorkshire, England, with an image size of $19008 \times 15296$. To show the separation degree of anomaly/normal regions, the binary detection maps were obtained with thresholds, ensuring the same anomaly recall rate of 70%. We use a yellow mask to represent the true anomalies and a red mask to represent the false alarms. In the first landslide event, the damaged area can be detected with few false alarms and basically matches the real region. Even though the second area has seven times the area of the first event, the fire area is detected successfully, with only four false alarms, which is acceptable in practical usage. The results obtained for the two events confirm that the proposed AnomalyCDM can effectively distinguish the normal and anomaly pixels in large-scale scenes.

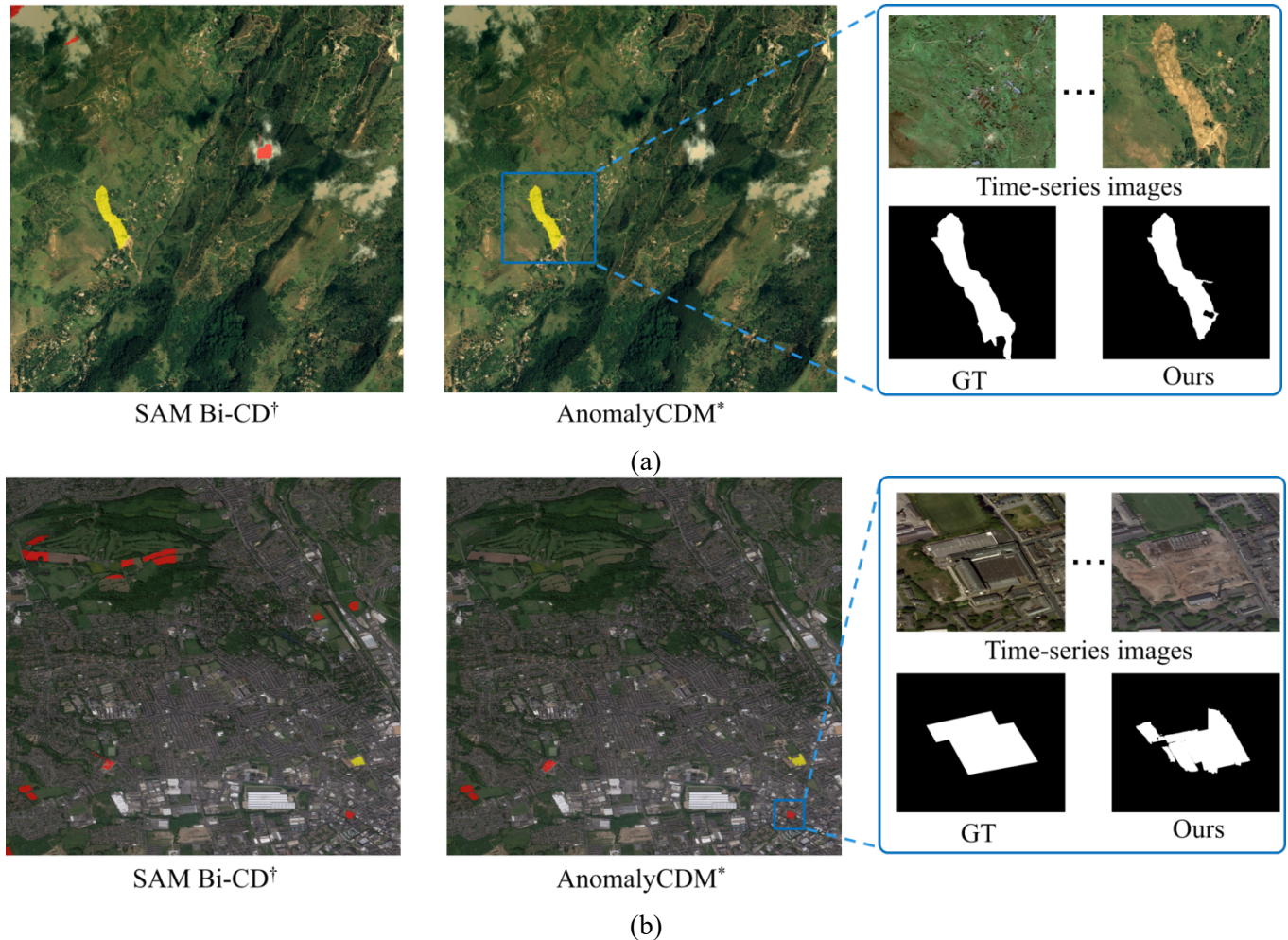

**Fig. 9.** Exemplified results of the proposed AnomalyCDM on two large-scale anomaly events. (a) The landslide event in Badul, Sri Lanka (2014) and (b) a fire event that broke out in West Yorkshire, England (2016). The yellow mask represents correct detections, and the red mask represents the false alarms. The results show that the proposed model has an accurate anomaly localization ability, even for large-scale and complex scenes.

### 5.2.3 Ablation analysis for the time-series observations

The prior comparison results indicated that the time-series observations can make the model aware of normal/anomalous changes. This section provides a detailed ablation analysis for each event. Fig. 10 reports the quantitative improvement of AnomalyCDM compared to the bi-temporal SAM Bi-CD model in the recall, precision, and $F_1$-score metrics. Each point represents an anomaly event, and the improvement amount shows the promotion after introducing the time-series observations. The results show that most events obtain a better performance with time-series images, compared to bi-temporal images.

To clarify the connection between the time step number and the detection performance, we filtered out

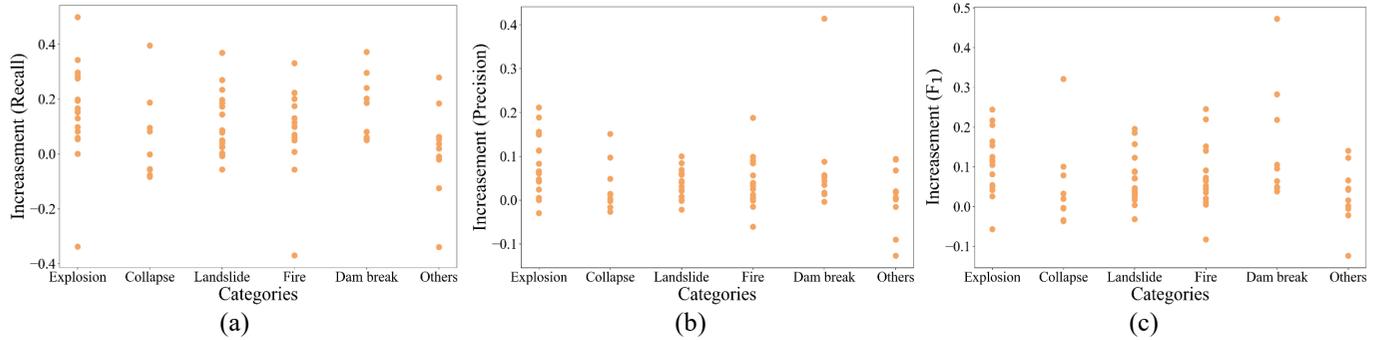

**Fig. 10.** The improvement of time-series anomaly change detection compared to the bi-temporal change detection results, which shows the event-level improvement in (a) recall, (b) precision, and (c) $F_1$-score. The results show that the use of the time-series information in the AnomalyCD technique can promote the performance.

39 anomaly events with at least four normal time steps and observed the performance change when reducing the time steps one by one. The results are reported in Table 4. For the 39 events, almost all the quantitative metrics become worse as the number of normal time steps is reduced, and the decrease amount is positively related to the number of time steps. After computing the average performance, we found that there appears to be a pattern where the $F_1$-score metric is reduced by roughly 2 points and the precision by 1 point on average for every reduction of normal time step.

**Table 4**

Quantitative results showing the connection between the time step number and detection performance, which show that the decreasing accuracy is positively related to the number of time steps. The colored numbers represent the difference in accuracy after reducing the normal time steps.

| Anomaly category | All the normal time steps | | | Normal time steps −1 | | | Normal time steps −2 | | | Normal time steps −3 | | |
|---|---|---|---|---|---|---|---|---|---|---|---|---|
| | R | P | F1 | R | P | F1 | R | P | F1 | R | P | F1 |
| Explosion | 74.78 | 47.81 | 58.33 | 72.69 | 47.48 | 57.44 | 68.15 | 45.29 | 54.42 | 58.61 | 42.09 | 48.99 |
| | | | | −2.09 | −0.33 | −0.89 | −6.63 | −2.52 | −3.91 | −16.17 | −5.72 | −9.34 |
| Collapse | 76.96 | 35.86 | 48.92 | 73.69 | 35.28 | 47.72 | 73.07 | 35.86 | 48.11 | 72.09 | 35.01 | 47.13 |
| | | | | −3.27 | −0.58 | −1.20 | −3.89 | 0.00 | −0.81 | −4.87 | −0.85 | −1.79 |
| Landslide | 77.36 | 54.64 | 64.05 | 74.89 | 54.23 | 62.90 | 67.43 | 52.69 | 59.16 | 72.89 | 54.68 | 62.48 |
| | | | | −2.47 | −0.41 | −1.15 | −9.93 | −1.95 | −4.89 | −4.47 | 0.04 | −1.57 |
| Fire | 64.29 | 67.27 | 65.75 | 63.30 | 66.75 | 64.98 | 56.09 | 65.62 | 60.48 | 54.33 | 64.66 | 59.04 |
| | | | | −0.99 | −0.52 | 0.77 | −8.20 | −1.65 | −5.27 | −9.96 | −2.61 | −6.71 |
| Dam break | 80.05 | 88.64 | 84.13 | 78.00 | 88.36 | 82.86 | 78.41 | 88.65 | 83.22 | 68.08 | 86.52 | 76.20 |
| | | | | −2.05 | −0.28 | −1.27 | −1.64 | 0.01 | −0.91 | −11.97 | −2.12 | −7.93 |
| Others | 40.38 | 33.82 | 36.81 | 36.26 | 31.86 | 33.92 | 32.49 | 30.81 | 31.62 | 35.48 | 32.15 | 33.73 |
| | | | | −4.12 | −1.96 | −2.89 | −7.89 | −3.01 | −5.19 | −4.90 | −1.67 | −3.08 |
| Average | 68.97 | 54.67 | 59.67 | 66.47 | 53.99 | 58.30 | 62.61 | 53.15 | 56.17 | 60.25 | 52.52 | 54.60 |
| | | | | −2.50 | −0.68 | −1.11 | −6.36 | −1.52 | −3.50 | −8.72 | −2.16 | −5.07 |

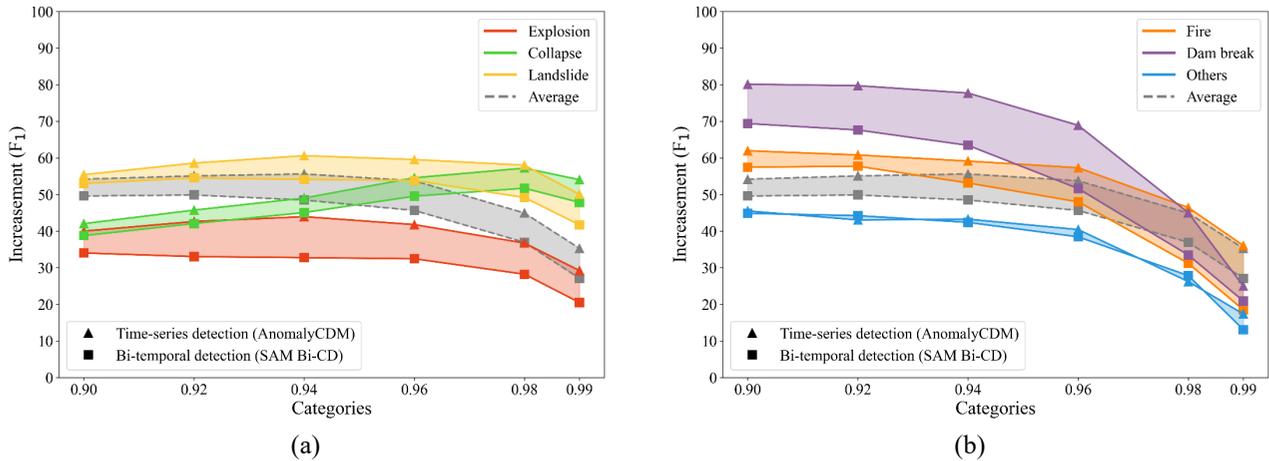

**Fig. 11.** Sensitivity analysis for the effect of the thresholding quantile on the detection performance (F1-score). With the increase of the quantile, the promotion of introducing time-series information also exists, despite the overall decreasing trend. By averaging all the anomaly categories (gray lines), the 0.94 quantile achieves the highest and most robust performance.

*5.2.4   Sensitivity analysis for the thresholding quantile*

Since there is no supervised training process, the zero-shot/unsupervised detection methods always output a continuous change density map, which requires a thresholding quantile to be converted into a binary map. The used 0.94 quantile was determined from the sensitivity analysis shown in Fig. 11. The quantile was varied from 0.90 to 0.99, corresponding to the statistical anomaly ratio in Section 3.2. With the increase of the quantile, it becomes more difficult to keep most anomaly regions detected, and the $F_1$-scores of most anomaly categories decrease. Despite the overall decreasing trend, the promotion of introducing time-series information also exists, which is shown as the colored regions in Fig. 11. By averaging all the anomaly categories (i.e., the gray area), we found that the proposed AnomalyCDM was robust to the threshold change before the 0.96 quantile, and the highest $F_1$-score was obtained with the 0.94 quantile.

*5.2.5   Sensitivity analysis for the inference patch size*

Each collected anomaly event in the AnomalyCDD dataset covers a large-scale scene, with the image width/height mostly larger than 10000 pixels. Limited by the memory usage, the original images have to be cropped into patches and inferred in a non-overlapped manner, where the cropped patch size is an important hyper-parameter. We conducted a sensitivity analysis for the patch size, and the results are listed in Table 5. We varied the patch size from 512 to 4096 pixels and recorded both the accuracy and average processing time, considering the efficiency. For the different anomaly categories, the optimal patch sizes are different when comparing the accuracy only. For example, a patch size of 1024 pixels is optimal for the explosion and collapse categories, while 4096 pixels is optimal for the landslide category. It can be deduced that landslide events always have a larger damaged area than explosion events, and a patch size of 1024 pixels may not provide the required contextual information. With the doubling of the patch size, the processing time is reduced by half. Processing a single anomaly event takes around 1 hour with a patch size of 512 pixels, but only 10 minutes with a patch size of 4096 pixels. Although the patch size of 4096 pixels has a faster processing speed and higher average $F_1$-score, we found that this setting performed worse in four categories than the size of 2048 pixels, and an improvement in average accuracy is only seen in the two categories of landslide and others. Thus, we finally chose a patch size of 2048 pixels, considering the performance on the different categories.

*5.2.6    Sensitivity analysis for the representation space*

The zero-shot detection performance of AnomalyCDM is supported by the general representation of the SAM foundation model, where the images are all mapped into the representation space and compared. In fact, AnomalyCDM is a framework that is compatible with different visual representations, and a more general

**Table 5**

Sensitivity analysis for the effect of the inference patch size on the detection performance.

| Anomaly category | 512 | | | 1024 | | | 2048 | | | 4096 | | |
|---|---|---|---|---|---|---|---|---|---|---|---|---|
| | R | P | F1 | R | P | F1 | R | P | F1 | R | P | F1 |
| Explosion | 49.81 | 34.12 | 40.50 | **60.38** | **38.74** | **47.20** | 55.68 | 36.30 | 43.95 | 52.06 | 37.77 | 43.78 |
| Collapse | 73.96 | **38.42** | 50.57 | **80.76** | 37.16 | **50.90** | 76.03 | 36.20 | 49.04 | 72.26 | 35.19 | 47.33 |
| Landslide | 62.04 | 53.91 | 57.69 | 71.28 | 56.18 | 62.83 | 72.51 | 52.07 | 60.61 | **76.89** | **58.89** | **66.70** |
| Fire | 55.15 | 49.38 | 52.11 | 59.30 | 53.06 | 56.01 | **64.96** | **54.26** | **59.13** | 57.83 | 53.60 | 55.64 |
| Dam break | 65.60 | 85.76 | 74.34 | 66.27 | 86.36 | 75.00 | **69.77** | 87.68 | **77.71** | 62.96 | **87.77** | 73.32 |
| Others | 27.84 | 35.70 | 31.29 | 38.71 | 42.42 | 40.48 | 42.43 | 44.14 | 43.27 | **49.39** | **46.29** | **47.79** |
| Average | 55.73 | 49.55 | 51.08 | 62.78 | 52.32 | 55.40 | **63.56** | 51.78 | 55.62 | 61.90 | **53.25** | **55.76** |
| Time | 3813.33 s | | | 2092.50 s | | | 1230.64 s | | | 553.77 s | | |

representation could further enhance the performance. In this section, we describe how we investigated the influence of different representation spaces, i.e., the Contrastive Language-Image Pre-Training (CLIP) vision-language model (Dong et al., 2024; Radford et al., 2021) and the SAM, which are two visual foundation models that have shown a state-of-the-art zero-shot ability in many domains. The new representation was used to discriminate the anomalous changes in the second stage. Two CLIP versions were tested with ResNet-50 and ResNet-101 backbones. The contextual information of segmented instances was also provided when using CLIP to obtain the embeddings. For the SAM, the effect of the representation space being processed by the neck module was also tested, which corresponded to two different representation spaces. The results are reported in Table 6. After removing the neck module in the SAM, the performance clearly drops (the average $F_1$-score is reduced from 55.62 to 25.54), showing the necessity of the neck module. The CLIP model with ResNet-50 as the backbone also shows a satisfactory performance, with an average $F_1$-score of 48.47. The lower performance may be caused by the different pre-training processes, where CLIP is trained for completing the image-level classification task, while the SAM is trained for the dense segmentation task. Compared to the ResNet-50 backbone, the ResNet-101 backbone performs unexpectedly worse, which seemly violates the scaling law. This is in fact caused by the property of the anomaly change detection task. With the

**Table 6**
Sensitivity analysis for the effect of different representation spaces on the detection performance. The latent space of the SAM with the neck module shows an obvious superiority.

| Anomaly category | CLIP-ResNet-50 | | | CLIP-ResNet-101 | | | SAM-w/o neck | | | SAM-w neck | | |
|---|---|---|---|---|---|---|---|---|---|---|---|---|
| | R | P | F1 | R | P | F1 | R | P | F1 | R | P | F1 |
| Explosion | 54.39 | **36.76** | 43.87 | 50.45 | 35.16 | 41.44 | 20.30 | 21.03 | 20.66 | **55.68** | 36.30 | **43.95** |
| Collapse | 64.32 | 33.33 | 43.91 | 66.04 | 32.69 | 43.74 | 33.46 | 12.92 | 18.64 | **76.03** | **36.20** | **49.04** |
| Landslide | 51.82 | 44.38 | 47.81 | 44.35 | 42.68 | 43.50 | 15.36 | 25.16 | 19.07 | **72.51** | **52.07** | **60.61** |
| Fire | 54.07 | 49.44 | 51.65 | 50.64 | 46.68 | 48.58 | 23.46 | 37.96 | 29.00 | **64.96** | **54.26** | **59.13** |
| Dam break | 57.11 | 83.43 | 67.80 | 54.82 | 83.19 | 66.09 | 29.60 | 69.59 | 41.53 | **69.77** | **87.68** | **77.71** |
| Others | 34.92 | 36.65 | 35.77 | 32.10 | 36.90 | 34.33 | 19.85 | 31.41 | 24.33 | **42.43** | **44.14** | **43.27** |
| Average | 52.77 | 47.33 | 48.47 | 49.73 | 46.22 | 46.28 | 23.67 | 33.01 | 25.54 | **63.56** | **51.78** | **55.62** |

**Table 7**

Sensitivity analysis of the effect of different distance metrics on the detection performance. From the perspective of average performance, no obvious difference can be observed, but the cosine distance metric shows a slightly better performance.

|  | Explosion ||| Collapse ||| Landslide ||| Fire ||| Dam break ||| Others ||| Average |||
|---|---|---|---|---|---|---|---|---|---|---|---|---|---|---|---|---|---|---|---|---|---|
|  | R | P | F1 | R | P | F1 | R | P | F1 | R | P | F1 | R | P | F1 | R | P | F1 | R | P | F1 |
| $L_2$ norm | 49.87 | 35.01 | 41.14 | 73.28 | 35.48 | 47.81 | **73.37** | 52.01 | **60.88** | **66.69** | 54.59 | **60.04** | 68.68 | 87.37 | 76.91 | 42.07 | **44.52** | 43.26 | 62.33 | 51.50 | 55.01 |
| $L_1$ norm | 49.61 | 34.99 | 41.04 | 73.00 | 35.29 | 47.58 | 72.93 | 52.02 | 60.72 | 65.28 | **54.77** | 59.56 | 69.28 | 87.57 | 77.36 | 42.11 | 43.81 | 42.94 | 62.04 | 51.41 | 54.87 |
| Cosine | **55.68** | **36.30** | **43.95** | **76.03** | **36.20** | **49.04** | 72.51 | **52.07** | 60.61 | 64.96 | 54.26 | 59.13 | **69.77** | **87.68** | **77.71** | **42.43** | 44.14 | **43.27** | **63.56** | **51.78** | **55.62** |

deepening of the network layers, the features extracted by ResNet-101 become more high level and contain more semantic information. However, the anomaly change detection task in this study is at a middle level, and too much semantic information can decrease the performance. Thus, AnomalyCDM finally uses the SAM representation space after the neck module in the second stage.

*5.2.7  Sensitivity analysis for the distance metric*

At both stages of AnomalyCDM, the visual representations were compared in the different time steps. In Section 5.2.6, we described how we tested the influence of the representation space. This section provides a sensitivity analysis of different distance metrics for the representation comparison. The results are provided in Table 7, where the classical $L_2$ norm, $L_1$ norm, and cosine distances were tested. From the perspective of the average accuracy, the influence of the different metrics is relatively small, compared to the different representation spaces. However, the cosine distance metric can bring an overall better performance, especially for the explosion anomaly category.

## 6.  Conclusion

As a new technique for the Earth anomaly detection task, AnomalyCD learns to distinguish anomalous changes through the use of historical monitoring images. Compared to the existing classification and change detection methods, AnomalyCD does not need human supervision and can be applied to anomalies with few samples or unknown anomalies. Compared to some of the unsupervised methods (Wu et al., 2023),

AnomalyCD has the ability to suppress the normal changes and reduce the false alarms. To benchmark the AnomalyCD technique, the anomaly change detection dataset (AnomalyCDD) was constructed with a high spatial resolution (0.15–2.39 m/pixel), time-series images (3–7 time steps), and large-scale (1927.93 km$^2$ in total) observations. As a baseline model for the built benchmark, AnomalyCDM was further developed, which can detect unseen and varied anomalies in a zero-shot manner with the extracted temporal representation from the SAM.

Despite the improvement, AnomalyCDM is a preliminary implementation of the AnomalyCD technique. For example, the normal changes are extracted and compared for each instance separately in AnomalyCDM, where the learned normal change patterns from different locations may also be helpful for the current instance, but are not utilized. Furthermore, a dynamic threshold rather than a fixed quantile will also be needed in a more powerful implementation.

**Acknowledgments**

This work was supported by the National Natural Science Foundation of China under Grant No. 42325105.

# References

Abera, T.A., Heiskanen, J., Maeda, E.E., Hailu, B.T., Pellikka, P.K.E., 2022. Improved detection of abrupt change in vegetation reveals dominant fractional woody cover decline in Eastern Africa. Remote Sens Environ 271, 112897.

Bovolo, F., Bruzzone, L., 2006. A theoretical framework for unsupervised change detection based on change vector analysis in the polar domain. IEEE Transactions on Geoscience and Remote Sensing 45, 218–236.

Brakhasi, F., Hajeb, M., Mielonen, T., Matkan, A., Verbesselt, J., 2021. Investigating aerosol vertical distribution using CALIPSO time series over the Middle East and North Africa (MENA), Europe, and India: A BFAST-based gradual and abrupt change detection. Remote Sens Environ 264, 112619.

Capliez, E., Ienco, D., Gaetano, R., Baghdadi, N., Salah, A.H., Le Goff, M., Chouteau, F., 2023. Multi-sensor temporal unsupervised domain adaptation for land cover mapping with spatial pseudo labelling and adversarial learning. IEEE Transactions on Geoscience and Remote Sensing.

Chen, H., Shi, Z., 2020. A spatial-temporal attention-based method and a new dataset for remote sensing image change detection. Remote Sens (Basel) 12, 1662.

Chen, H., Song, J., Yokoya, N., 2024a. Change Detection Between Optical Remote Sensing Imagery and Map Data via Segment Anything Model (SAM).

Chen, H., Song, J., Yokoya, N., 2024b. Change Detection Between Optical Remote Sensing Imagery and Map Data via Segment Anything Model (SAM). arXiv preprint arXiv:2401.09019.

CRED / UCLouvain. EM-DAT: The International Disaster Database [Internet]. Brussels, Belgium: Centre for Research on the Epidemiology of Disasters; [cited 2024 Aug 8]. Available from: http://www.emdat.be/.

Dong, S., Wang, L., Du, B., Meng, X., 2024. ChangeCLIP: Remote sensing change detection with multimodal vision-language representation learning. ISPRS Journal of Photogrammetry and Remote Sensing 208, 53–69.

He, Y., Wang, J., Zhang, Y., Liao, C., 2024a. An efficient urban flood mapping framework towards disaster response driven by weakly supervised semantic segmentation with decoupled training samples. ISPRS Journal of Photogrammetry and Remote Sensing 207, 338–358.

He, Y., Wang, J., Zhang, Y., Liao, C., 2024b. An efficient urban flood mapping framework towards disaster response driven by weakly supervised semantic segmentation with decoupled training samples. ISPRS Journal of Photogrammetry and Remote Sensing 207, 338–358.

Ji, S., Wei, S., Lu, M., 2018. Fully convolutional networks for multisource building extraction from an open aerial and satellite imagery data set. IEEE Transactions on geoscience and remote sensing 57, 574–586.

Khandelwal, A., Karpatne, A., Marlier, M.E., Kim, J., Lettenmaier, D.P., Kumar, V., 2017. An approach for global monitoring of surface water extent variations in reservoirs using MODIS data. Remote Sens Environ 202, 113–128.

Kirillov, A., Mintun, E., Ravi, N., Mao, H., Rolland, C., Gustafson, L., Xiao, T., Whitehead, S., Berg, A.C.,

Lo, W.-Y., 2023. Segment anything, in: Proceedings of the IEEE/CVF International Conference on Computer Vision. pp. 4015–4026.

Lebedev, M.A., Vizilter, Y. V, Vygolov, O. V, Knyaz, V.A., Rubis, A.Y., 2018. Change detection in remote sensing images using conditional adversarial networks. The International Archives of the Photogrammetry, Remote Sensing and Spatial Information Sciences 42, 565–571.

Li, J., Li, Z.-L., Wu, H., You, N., 2022. Trend, seasonality, and abrupt change detection method for land surface temperature time-series analysis: Evaluation and improvement. Remote Sens Environ 280, 113222.

Lipton, A.J., Fujiyoshi, H., Patil, R.S., 1998. Moving target classification and tracking from real-time video, in: Proceedings Fourth IEEE Workshop on Applications of Computer Vision. WACV'98 (Cat. No. 98EX201). IEEE, pp. 8–14.

Merz, B., Blöschl, G., Vorogushyn, S., Dottori, F., Aerts, J.C.J.H., Bates, P., Bertola, M., Kemter, M., Kreibich, H., Lall, U., 2021. Causes, impacts and patterns of disastrous river floods. Nat Rev Earth Environ 2, 592–609.

Pang, G., Shen, C., Cao, L., Hengel, A. Van Den, 2021. Deep learning for anomaly detection: A review. ACM computing surveys (CSUR) 54, 1–38.

Radford, A., Kim, J.W., Hallacy, C., Ramesh, A., Goh, G., Agarwal, S., Sastry, G., Askell, A., Mishkin, P., Clark, J., 2021. Learning transferable visual models from natural language supervision, in: International Conference on Machine Learning. PMLR, pp. 8748–8763.

Tian, S., Ma, A., Zheng, Z., Zhong, Y., 2020. Hi-UCD: A Large-scale Dataset for Urban Semantic Change Detection in Remote Sensing Imagery.

Toker, A., Kondmann, L., Weber, M., Eisenberger, M., Camero, A., Hu, J., Hoderlein, A.P., Şenaras, Ç., Davis, T., Cremers, D., 2022. DynamicEarthNet: Daily multi-spectral satellite dataset for semantic change segmentation, in: Proceedings of the IEEE/CVF Conference on Computer Vision and Pattern Recognition. pp. 21158–21167.

Tupper, A.C., Fearnley, C.J., 2023. Disaster early-warning systems can succeed—but collective action is needed. Nature 623, 478–482.

Wei, H., Jia, K., Wang, Q., Cao, B., Qi, J., Zhao, W., Yang, J., 2023. Real-time remote sensing detection framework of the earth's surface anomalies based on a priori knowledge base. International Journal of Applied Earth Observation and Geoinformation 122, 103429.

Wu, C., Du, B., Zhang, L., 2023. Fully convolutional change detection framework with generative adversarial network for unsupervised, weakly supervised and regional supervised change detection. IEEE Trans Pattern Anal Mach Intell.

Xian, G.Z., Smith, K., Wellington, D., Horton, J., Zhou, Q., Li, C., Auch, R., Brown, J.F., Zhu, Z., Reker, R.R., 2022. Implementation of the CCDC algorithm to produce the LCMAP Collection 1.0 annual land surface change product. Earth Syst Sci Data 14, 143–162.

Yang, K., Xia, G.-S., Liu, Z., Du, B., Yang, W., Pelillo, M., Zhang, L., 2021. Asymmetric Siamese networks for semantic change detection in aerial images. IEEE Transactions on Geoscience and Remote Sensing 60, 1–18.

Yar, H., Ullah, W., Khan, Z.A., Baik, S.W., 2023. An Effective Attention-based CNN Model for Fire Detection in Adverse Weather Conditions. ISPRS Journal of Photogrammetry and Remote Sensing 206, 335–346.

Zhang, X., Yu, W., Pun, M.-O., Shi, W., 2023. Cross-domain landslide mapping from large-scale remote sensing images using prototype-guided domain-aware progressive representation learning. ISPRS Journal of Photogrammetry and Remote Sensing 197, 1–17.

Zheng, Z., Zhong, Y., Wang, J., Ma, A., Zhang, L., 2021. Building damage assessment for rapid disaster response with a deep object-based semantic change detection framework: From natural disasters to man-made disasters. Remote Sens Environ 265, 112636.

Zheng, Z., Zhong, Y., Zhang, L., Ermon, S., 2024. Segment Any Change. arXiv preprint arXiv:2402.01188.

Zhu, Z., Woodcock, C.E., 2014. Continuous change detection and classification of land cover using all available Landsat data. Remote Sens Environ 144, 152–171.